\documentclass[lettersize,journal]{IEEEtran}
\IEEEoverridecommandlockouts
\usepackage{tabularx}
\usepackage[T1]{fontenc}% optional T1 font encoding
\usepackage{algorithm}
\usepackage[noend]{algpseudocode}
\usepackage{graphicx}
\usepackage{amsmath,amssymb,amsfonts,stfloats}
\usepackage{array}
\usepackage{textcomp}
\usepackage[font=small]{caption}
\usepackage{paralist}
\usepackage{lettrine}
\usepackage{cite}
\usepackage{xcolor}
\usepackage{cancel}
\usepackage{booktabs}
\usepackage{mathtools}
\usepackage{bbm}
\usepackage{array}
\usepackage{booktabs}
\usepackage[caption=false,font=footnotesize]{subfig}
\usepackage{url}
\usepackage{acronym}
\usepackage{multicol}
\usepackage{makecell}
\usepackage{enumitem}
\usepackage{bm}
\usepackage[pagewise]{lineno}
\usepackage{multirow} % Required for multirow

\begin{document}
%%
% \pagewiselinenumbers   
% \switchlinenumbers  
\title{Pilot Contamination-Aware Graph Attention Network for Power Control in CFmMIMO}

\author{Tingting Zhang,~\IEEEmembership{Student Member,~IEEE}, Sergiy A. Vorobyov,~\IEEEmembership{Fellow,~IEEE},
David J. Love,~\IEEEmembership{Fellow,~IEEE}, \\
Taejoon Kim,~\IEEEmembership{ Senior Member,~IEEE}, and Kai Dong,~\IEEEmembership{ Member,~IEEE}
%\author{Tingting Zhang,~\IEEEmembership{Student member,~IEEE,}
        % <-this % stops a space
\thanks{This work was supported in part by the Academy of Finland under Grant 357715.}% <-this % stops a space
\thanks{Tingting Zhang and Sergiy A. Vorobyov are  with the Department of Information and Communications Engineering, Aalto University, 02150 Espoo, Finland (e-mail: tingting.zhang@aalto.fi, sergiy.vorobyov@aalto.fi).}
\thanks{David J. Love is with the School of Electrical and Computer Engineering, Purdue University, West Lafayette, IN 47907 USA (e-mail:
djlove@purdue.edu).}
\thanks{Taejoon Kim is with the School of Electrical, Computer and Energy Engineering, Arizona State University, Tempe, AZ 85287 USA (e-mail:
taejoonkim@asu.edu).}
\thanks{Kai Dong is with the Center for Wireless Communications, University of
Oulu, 90570 Oulu, Finland (e-mail: kai.dong@oulu.fi).}
}

\maketitle

\begin{abstract}
Optimization-based power control algorithms are predominantly iterative with high computational complexity, making them impractical for real-time  applications in cell-free massive multiple-input multiple-output (CFmMIMO) systems. Learning-based methods have emerged as a promising alternative, and among them, graph neural networks (GNNs) have demonstrated their excellent performance in solving power control problems. 
However, all existing GNN-based approaches assume ideal orthogonality among pilot sequences for user equipments (UEs),  which is unrealistic given that the number of UEs exceeds the available orthogonal pilot sequences in CFmMIMO schemes.  
% Moreover, most learning-based methods assume a fixed number of UEs, whereas the number of active UEs varies over time in practice.
Additionally,  supervised training necessitates costly computational resources for computing the target power control solutions for a large volume of training samples.
To address these issues, we propose a graph attention  network for downlink power control in CFmMIMO systems that operates in a self-supervised manner while effectively handling pilot contamination and adapting to a dynamic number of UEs. Experimental results show its effectiveness, even in comparison to the optimal accelerated projected gradient method as a baseline.
\end{abstract}


\section{Introduction}
Cellular massive multiple-input multiple-output (MIMO), usually enabled by multi-antenna panel arrays, is the backbone of today's 5G communication technology. However, a persistent problem with cellular massive MIMO deployments is the poor service quality experienced by the cell-edge user equipments (UEs).
\makebox{Cell-free} massive MIMO (CFmMIMO) has been proposed as a promising architecture for next-generation wireless communications.
CFmMIMO eliminates the cellular boundaries  and  instead deploys massive distributed access points (APs) across a large coverage area \cite{b_CFm}. These APs cooperatively serve UEs within the coverage area without the constraints of cell  boundaries, ensuring fairness in service quality across UEs at different locations and reducing interference through efficient beamforming enabled by the coordination of a massive number of distributed antennas. 
As a result, CFmMIMO can achieve a higher network spectral efficiency (SE) compared to  traditional cellular schemes \makebox{\cite{b14, bDJ}}.

To fully realize the benefits of CFmMIMO, several challenges must be addressed. In particular,  power control\footnote{We focus on downlink power control in a centralized fashion in this paper.} and pilot allocation are critical.
With an increasing number of UEs,  especially in CFmMIMO systems, the limited number of orthogonal pilot sequences becomes insufficient, leading to pilot contamination.  Pilot contamination significantly degrades the communication quality, and thus,  its impact cannot be overlooked when addressing  power control problems.  % !
Traditional power control algorithms are predominantly optimization-based, with computational complexity increasing sharply as the numbers of APs and UEs grow.
For CFmMIMO systems with a large number of APs and UEs, existing algorithms lead to high computational costs and processing time, which runs counter to the demands for high efficiency and low latency.

To address this challenge, learning-based algorithms have been proposed to compute power allocation coefficients with significantly improved efficiency.  Various neural network models, including convolutional neural networks \cite{b3, b33},  transformer neural networks \cite{b9, b12},  and graph neural networks (GNNs) \cite{b5, b6,b7, b8, b88, b+R} have been explored. %using supervised or self-supervised training approaches. 
GNNs have demonstrated  excellent performance in solving power control problems thanks to their  structural similarity to wireless communication networks \cite{b6, b7}.
As mentioned earlier,  pilot contamination cannot be ignored when addressing power control. However, to the best of our knowledge, all existing \makebox{GNN-based} methods assume the use of ideally orthogonal pilot sequences, which is a big stretch, especially in CFmMIMO systems. 
%In addition, most \makebox{learning-based} methods assume CFmMIMO systems with a fixed number of UEs \cite{b5,b2, b4,b+}, which is impractical given the dynamic nature of real-world scenarios. 
To bridge this  gap, we propose a pilot contamination-aware \makebox{graph} attention network to handle power control in CFmMIMO systems.
The proposed network model utilizes an  attention mechanism \cite{b9, b12} enhanced GNN to achieve power control with the awareness of pilot contamination in  dynamic scenarios with a varying number of UEs. % without  requiring retraining
Besides, the network is trained in a self-supervised manner,  reducing the workload of computing target power control coefficients.
Experimental results are presented to compare the proposed method against  the first-order  accelerated  projected gradient (APG) method described in \cite{b10} and transformer in \cite{b12}. 

\section{System Model}
\subsection{Cell-Free Massive MIMO}
Consider the downlink power control  in a CFmMIMO system, where all UEs are served by APs using common \makebox{time-frequency} resources, and all APs are synchronized.
Stringent synchronization among UEs and APs in both time and frequency domains is essential to enable coherent joint transmission in CFmMIMO architectures. Achieving such synchronization is technically challenging, especially at large scale and with distributed APs, and requires dedicated algorithms and network infrastructure.
Several practical approaches have been developed to address this challenge. In cell-free architectures, synchronization is often maintained via a combination of centralized processing at the control unit, high-capacity fronthaul links, and regular synchronization signals exchanged between the APs and the central unit. Solutions can include over-the-air synchronization procedures, network-based protocols, and periodic calibration, sometimes leveraging GPS or precise network timing protocols (e.g., IEEE 1588  Precision Time Protocol). Experimental and field trials have demonstrated the feasibility of achieving tight synchronization in such distributed settings.
Further advancements on this topic in terms of scalability and cost are still required \cite{b_CFm}.

The system operates in time-division duplex (TDD) mode. 
The system consists of $M$ APs and $K$ \makebox{single-antenna} UEs  randomly distributed over a large coverage area, with each AP equipped with $N$ antennas. 
Assuming that channels are block-fading and remain constant over a coherence block of  $T_{\rm c}$ symbols,  the uplink channel from the $k$th  UE to the $m$th AP can be expressed as
\begin{equation}
  \begin{aligned}
\mathbf{g}_{m,k} =  \sqrt{\beta_{m,k}} {\mathbf{h}_{m,k}},
\end{aligned}
\end{equation}
where $\beta_{m,k}$ denotes the large-scale fading coefficient, and $\mathbf{h}_{m,k} \in \mathbb{C}^N$ denotes the small-scale coefficient vector for  an \makebox{$N$-antenna} array. The entries of $\mathbf{h}_{m,k}$ are independent and identically distributed (i.i.d.) complex Gaussian random variables with zero mean and unit variance, i.e., $\mathcal{CN}(0,1)$.

 Pilot sequences  consisting of $T_{\rm p}$ ($T_{\rm p}\ll T_{\rm c} $) symbols  are  used for channel estimation during the uplink training stage. Channel reciprocity is assumed, enabling  downlink channel estimation using   uplink pilot sequences.
  Let the pilot sequence for the $k$th UE be $\sqrt{T_{\rm p}} \boldsymbol{\varphi}_k \in \mathbb{C}^{T_{\rm p}}$ with $\|\boldsymbol{\varphi}_k\|^2=1$, where $\|\cdot\|$ represents the Euclidean norm of a vector. 
  In CFmMIMO systems, where a large number of UEs are served simultaneously, the condition $T_{\rm p}<K$ often arises  due to the limited length of pilot sequences, resulting in pilot contamination.
%Addressing pilot contamination within learning-based approaches is a key challenge considered in this paper.

  In the  uplink training phase,  the received signal at the $m$th AP  can be expressed as 
  \begin{equation}
  \begin{aligned}
\mathbf{Y}_{m}^{\rm up} =  \sqrt{\zeta_{\rm p} T_{\rm p}} \sum_{i=1}^K \mathbf{g}_{m,i} \boldsymbol{\varphi}_i^{\rm H} + \mathbf{W}_{m}^{\rm up},
\end{aligned}
\end{equation}
where  $\zeta_{\rm p}$ denotes  the normalized transmit  signal-to-noise ratio  per pilot symbol, $\left(\cdot\right)^{\rm H}$ represents the conjugate transpose operation, and $\mathbf{W}_{m}^{\rm up} \in \mathbb{C}^{N \times T_{\rm p}}$  is  the noise matrix whose entries follow i.i.d. $\mathcal{CN}(0,1)$. The minimum mean square error (MMSE) estimate of  the channel between the $m$th AP and the $k$th UE can be derived as \cite{b10}
\begin{equation}\label{3}
  \begin{aligned}
\hat{\mathbf{g}}_{m, k}=\frac{\sqrt{\zeta_{\rm p} T_{\rm p}} \beta_{m, k}}{1+\zeta_{\rm p}  T_{\rm p} \sum_{i=1}^K \beta_{m, i}\left|\boldsymbol{\varphi}_i^{\rm H} \boldsymbol{\varphi}_k\right|^2} \mathbf{Y}_{m}^{\rm up} \boldsymbol{\varphi}_k,
\end{aligned}
\end{equation}
where  $\left| \cdot \right|$ denotes the absolute value. The mean square  of each entry  of $\hat{\mathbf{g}}_{m, k}$ is 
\begin{equation}
  \begin{aligned}
\overline{g}_{m, k}=\mathbb{E}\left[\left|\hat{\mathbf{g}}_{m, k}[n]\right|^2\right]=\frac{\zeta_{\rm p}  T_{\rm p} \beta_{m, k}^2}{1+\zeta_{\rm p}  T_{\rm p}  \sum_{i=1}^K \beta_{m, i}\left|\boldsymbol{\varphi}_i^{\rm H} \boldsymbol{\varphi}_k\right|^2}.
\end{aligned}
\end{equation}

APs  cooperate to deal with  power control using large-scale fading information, which is assumed to be available through the central  processing unit.
Accurate large-scale fading information is critical for power allocation in CFmMIMO systems.
In practice, large-scale fading coefficients vary slowly over time when UEs are stationary or moving at slow speeds, so the channel can be modeled as block-fading.
These coefficients can be obtained through channel estimation using periodic pilot transmission and uplink-downlink reciprocity in TDD systems.
For higher-mobility UEs, one can employ tracking algorithms (e.g., Kalman filters) along with robust estimation techniques over successive coherence intervals to track the variations of large-scale fading \cite{Liu_Radar,Jayaprakasam_Robust}, thus providing real‑time large‑scale fading estimates.
The use of small-scale channel coefficients for power control is constrained by two key factors.
First, small-scale fading varies rapidly over time, requiring frequent updates. Relying on it would impose significant  computational overhead, making it impractical for large-scale network-level optimization tasks.
Second, unlike beamforming, power control is primarily influenced by the large-scale channel characteristics;
the impact of small-scale fading on power control solutions is relatively limited.
Therefore, even though the small-scale channel estimates are available,  only the large-scale information is used in most CFmMIMO power control \cite{Ghazanfari_Enhanced, Van_Large, Chen_Power}.

Denote the power control coefficient matrix as $\mathbf{M} \in \mathbb{R}^{M \times K}$, and the  $(m, k)$-th matrix element, $\mu_{m, k}$, is the power control coefficient for the signal transmitted from the $m$th AP to the $k$th UE.
According to the practical constraint on the total transmit power per AP, the feasible set of power control coefficients can be expressed as
\begin{equation}\label{cons}
  \begin{aligned}
  \mathcal{S}=
  \left\{\mathbf{M} \mid \mu_{m, k} \geq 0 ;\left\|\bm{\mu}_m \right\|^2 \leq \frac{1}{N},  \forall m,k\right\},
\end{aligned}
\end{equation}
where $\bm{\mu}_m$ is the $m$th row of $\mathbf{M}$.
Denoting the data symbol transmitted to the $k$th UE as $s_k$ with $\mathbb{E}\left[\left|s_k\right|^2\right]=1$,  and $\zeta _{\rm d}$ as the maximum  downlink transmit power per symbol  normalized to noise power at each AP,  the downlink  signal  from the $m$th AP using conjugate beamforming  can be expressed as
\begin{equation} \label{e1}
  \begin{aligned}
  \mathbf{x}_m=\sqrt{\zeta_{\rm d}} \sum_{k=1}^K \frac{\mu_{m, k}}{\sqrt{\overline{g}_{m, k}}}\hat{\mathbf{g}}_{m, k}^* s_k,
\end{aligned}
\end{equation}
where $\left(\cdot\right)^*$ denotes conjugate operation.
%It can be seen from \eqref{e1} that  the transmi power at the $m$th AP is  $\mathbb{E}\left[\left\|\mathbf{x}_m\right\|^2\right]=\zeta_{\rm d} N \left\|\boldsymbol{\mu}_m\right\|^2$. 
Using the use-and-then-forget bounding technique, the signal-to-interference-plus-noise ratio for the $k$th UE can be derived as \cite{b10}
\begin{equation} \label{gamma}
  \begin{aligned}
\gamma_k=  \frac{\zeta_{\rm d} \left(\overline{\bm{\mu}}_k^{\rm T} \boldsymbol{\nu}_{k, k}\right)^2}{\sum_{\substack{i=1 \\ i \neq k}}^K \zeta_{\rm d} \left(\overline{\bm{\mu}}_i^{\rm T} \boldsymbol{\nu}_{i, k}\right)^2+\frac{\zeta_{\rm d}}{N} \sum_{i=1}^K \left\|\overline{\mathbf{B}}_k \overline{\bm{\mu}}_i\right\|^2+\frac{1}{N^2}},
\end{aligned}
\end{equation}
where  $\overline{\bm{\mu}}_k$ denotes the $k$th column of  $\mathbf{M}$, 
$(\cdot)^{\rm T}$ is  the transpose operator,
$\overline{\mathbf{B}}_k $  is a diagonal matrix with  diagonal elements $\sqrt{\beta_{1 k}}, \ldots, \sqrt{\beta_{M k}}$, and the $m$th element of  $\boldsymbol{\nu}_{i k}\in \mathbb{R}^M$  is
\begin{equation} 
  \begin{aligned}
  \boldsymbol{\nu}_{i, k}[m]=\left|\boldsymbol{\varphi}_i^{\rm H} \boldsymbol{\varphi}_k\right| \sqrt{\overline{g}_{m, i}} \frac{\beta_{m, k}}{\beta_{m, i}}.
  \end{aligned}
\end{equation}
The downlink network SE of the $k$th UE is then 
\begin{equation} \label{se}
  \begin{aligned}
\mathrm{SE}_k=\left(1-\frac{T_{\rm p}}{T_{\rm c}}\right) \log_2 \left(1+\gamma_k\right).
  \end{aligned}
\end{equation}

\subsection{Optimization Problem Formulation}
Define $\mathbf{B} \in \mathbb{R}_+^{M \times K}$ as  the large-scale fading coefficient matrix with $\beta_{m, k}$ as the $(m, k)$-th element,  and  let $\boldsymbol{\Phi} \in \mathbb{R} ^{K\times K}$ denote the pilot allocation matrix, where the $(i, j)$-th element is given by $\left|\boldsymbol{\varphi}_i^{\rm H}  \boldsymbol{\varphi}_j\right|$. Here $\mathbb{R}_+^{M \times K}$ represents the set of matrices of dimension $M \times K$ with non-negative real entries.
From \eqref{gamma} and \eqref{se},  SE depends on both  the power control coefficients $\mathbf{M}$  and pilot allocation matrix $\boldsymbol{\Phi}$,  given a large-scale fading matrix $\mathbf{B}$. When $T_{\rm p}\geq K$, pilot sequences are  orthogonal across all UEs, making their impact on SE negligible. 
However, in massive MIMO scenarios, $T_{\rm p} < K$ often occurs.
Therefore, incorporating the information of pilot assignment is essential for optimizing power control under pilot contamination.

Max-min fairness maximization is the objective given by
\begin{equation}\label{ob}
    \begin{aligned} 
    \begin{array}{rrr}
\underset{\mathbf{M}}{\operatorname{maximize}}   \left\{ \underset{1 \leq k  \leq K}{\operatorname{min}} \mathrm{SE}_{k}(\mathbf{M} ; \boldsymbol{\Phi}, \mathbf{B}) \right\}.
\end{array}
    \end{aligned} 
\end{equation}
By adopting a smoothing technique with smoothing parameter $\lambda$,  the objective function in \eqref{ob} can be reformulated as  \cite{b10, b10+}
\begin{equation} \label{u}
  \begin{aligned}
  u(\mathbf{M} ; \boldsymbol{\Phi}, \mathbf{B})=-\frac{1}{\lambda} \log \left(\frac{1}{K} \sum_{k=1}^K \exp\left({-\lambda \, \mathrm{SE}_k(\mathbf{M} ; \boldsymbol{\Phi}, \mathbf{B})}\right) \right).
  \end{aligned}
\end{equation}

Define a deep neural network as a map, $\mathbf{M} =f(\boldsymbol{\Phi}, \mathbf{B}; \mathbf{\Theta})$. Here $f$ maps $\boldsymbol{\Phi}$ and $\mathbf{B}$ to the power control matrix $\mathbf{M}$, and $\mathbf{\Theta}$ represents the trainable weights of the network. 
Our goal is to determine the optimal weights $\mathbf{\Theta}_{\rm opt}$ using a self-supervised learning method, in which the neural network is trained on unlabeled data, thus eliminating the computational burden of generating target power control solutions, for example, via conventional optimization algorithms.
In this framework,  the network learns the mapping between its input and power control coefficients  $\mathbf{M}$ with the aim of maximizing the objective function in \eqref{u}.

\section{Graph Attention Network for Power Control}
\subsection{Graph Representation}
Consider a heterogeneous graph $\mathcal{G}=(\mathcal{V}, \mathcal{E})$ composed of a set of nodes $\mathcal{V}$ and edges $\mathcal{E}$. Each node $v$ represents an AP-UE pair and is mapped by $\psi(m, k)=v$, where $\psi$ is the mapping function from an AP-UE pair to its corresponding node, and $v=M(k-1)+m$.
% To enable a single power control network capable of handling a varying number of UEs, $K$ is set as the maximum number of UEs, $K_{\rm max}$, within the coverage area. If the actual number of  UEs, $K_{\rm act}$, is smaller than $K_{\rm max}$, an extra $K_{\rm max}-K_{\rm act}$ UEs are added as padding. To differentiate between actual and padded UEs, each node is assigned a binary attribute, ${\rm Mask}$, and  ${\rm Mask}_i=1$ indicates that the UE  associated with the $i$th node is served, while  ${\rm Mask}_i=0$ denotes a padded UE.
The bidirectional edge between the $i$th and $j$th nodes is denoted as $e_{i, j}$ for $i \neq j$. Edges are categorized based on whether the connected nodes share the same AP or UE, or neither:
\begin{itemize}
    \item \textbf{AP-type edges ($e_{i, j}^{\rm AP}$)}: Nodes share the same AP.
    \item \textbf{UE-type edges ($e_{i, j}^{\rm UE}$)}: Nodes share the same UE.
    \item \textbf{No connection}: If two nodes share neither an AP nor a UE, no edge exists between them.
\end{itemize}

\begin{figure}[t]
  \centerline{\includegraphics[width=7cm,height=4.7cm]{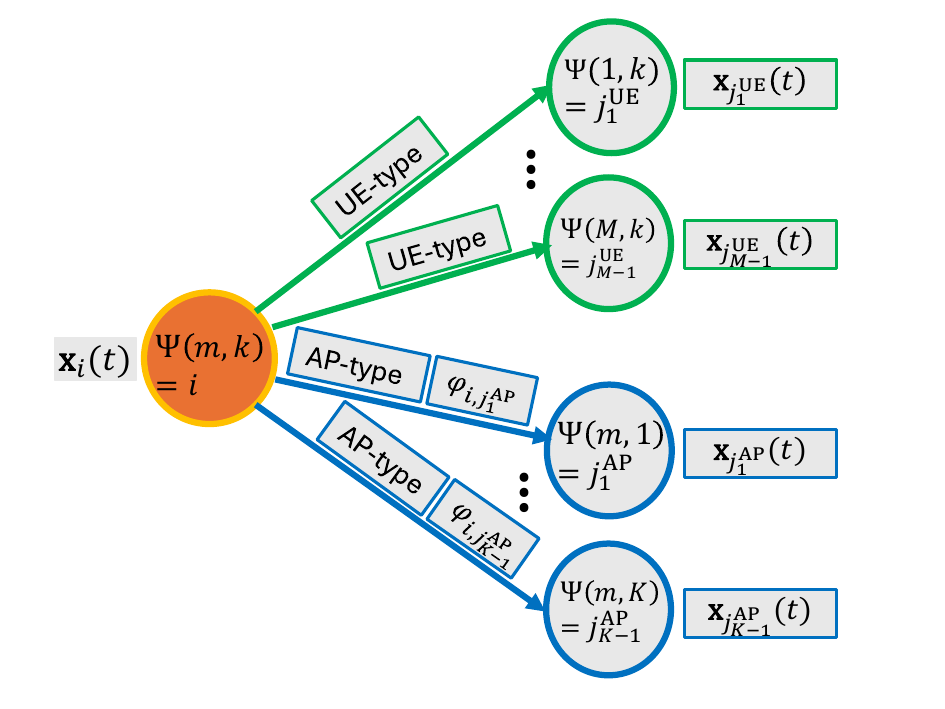}}
%  \vspace{1.5cm}
\caption{Illustration of graph characteristics.}
\label{graph}
\end{figure}

To incorporate pilot contamination awareness,  the pilot allocation matrix $\boldsymbol{\Phi}$ is included as an attribute for AP-type edges, capturing the orthogonality of allocated pilot sequences among UEs that share the same APs.
Specifically,  the \makebox{$(k_1, k_2)$-th} element of $\boldsymbol{\Phi}$, denoted by $\overline{\varphi}_{k_1, k_2}$, is assigned to those \makebox{AP-type} edges connecting nodes corresponding to the $k_1$-th and \makebox{$k_2$-th} UEs. 
To align with the GNN framework, $\varphi_{i, j}=\overline{\varphi}_{k_1, k_2}$ is introduced to represent the contamination information between nodes $i$ and $j$, which correspond to  $k_1$-th and $k_2$-th UEs, respectively.
% Specifically,  the \makebox{$(i, j)$-th} element of $\boldsymbol{\Phi}$, denoted as $\varphi_{i j}$, is assigned to those AP-type edges whose two connected nodes correspond to the $i$th and $j$th UEs. 
Suppose there are $T$ attention layers in the graph structural network, and the input and output features of the $i$th node at the $t$th graph attention layer are represented by $\mathbf{x}_i(t-1)$ and $\mathbf{x}_i(t)$, respectively, where $t=1,\ldots, T$. 
The characteristics of the proposed graph structure, including the node representation $\mathbf{x}_i(t)$, % the node attribute ${\rm Mask}_i$, 
the \makebox{AP-type} edges with $\varphi_{i, j}$, and the \makebox{UE-type} edges,  are illustrated in \makebox{Fig. \ref{graph}}. This structure enables the  network to effectively model the relationship between APs and UEs while accounting for pilot contamination.

\subsection{The Proposed  Network Model}
\textit{1) Preprocessing Stage:}
% To accommodate a varying number of UEs, the inputs $\mathbf{B}$ and $\boldsymbol{\Phi}$ are first padded. Specifically,  zero-padding is applied to generate  $\mathbf{B}_{\rm padded}$   of size $M \times K_{\rm max}$ and $\boldsymbol{\Phi}_{\rm padded}$ of size $K_{\rm max} \times K_{\rm max}$. For simplicity of notation,  $\mathbf{B}$  and $\boldsymbol{\Phi}$ are used to refer to their padded counterparts, and $K=K_{\rm max}$   in the following discussion unless explicitly stated.
Given that the values of large-scale fading coefficients span different orders of magnitude, the natural logarithm operation is applied to  each element of  $\mathbf{B}$, resulting in  $\mathbf{B}^{\rm log}$. Following this,  $\mathbf{B}^{\rm log}$ is transformed to have zero mean  and unit variance across all elements, resulting in $\mathbf{B}^{\rm std}$. 
Next, scaling and shifting are performed on  $\mathbf{B}^{\rm std}$ along each column, which can be expressed as
\begin{equation} 
  \begin{aligned}
  \mathbf{B}^{\rm pre} =\left( \boldsymbol{\alpha} \mathbf{1}_K ^{\rm T}\right) * \mathbf{B}^{\rm std} + \boldsymbol{\beta} \mathbf{1}_K ^{\rm T},
     \end{aligned}
\end{equation} 
where $\boldsymbol{\alpha}$ and $\boldsymbol{\beta}$  represent the trainable parameters for scaling and shifting, respectively, $\mathbf{1}_K$ is an all-ones vector of length $K$, and $*$ denotes the Hadamard product of matrices. Finally, $\mathbf{B}_{\rm pre}$  is vectorized into $\mathbf{b} \in \mathbb{R}^{MK\times 1}$. The $i$th element of $\mathbf{b}$ is taken as the initial representation of the $i$th node, also denoted as $\mathbf{x}_i(0)$ in the context of  the attention mechanism block.

\textit{2) Attention Mechanism Block:}
The attention mechanism allows the model to focus on specific parts of the input data by assigning varying importance to different components.
To capture the interrelationship among nodes, a  single-head attention mechanism is introduced.
%where masking enables the network to adapt to scenarios where the number of UEs varies. 
The attention mechanism block comprises $T = 4$ hidden layers, with  $\mathbf{x}_i(0)$ as the input to the first attention layer, the output sizes of  $4$  layers are $32,64, 64, 64$, respectively. A detailed explanation of the attention mechanism block  is provided in \makebox{Subsection \ref{atten}}.

\textit{3) Postprocessing Stage:} Given the attention mechanism block's output ${\mathbf X}(4) \in \mathbb{R}^{MK \times 64}$, whose $i$th row  corresponds to  $\mathbf{x}_i (4)$,  a combination of linear and nonlinear processing is first applied. The resulting output can be described as
\begin{equation} 
  \begin{aligned}
{\mathbf Y}_1= {\mathcal L}_2 ({\rm ReLU}({\mathcal L}_1 ({\mathbf X}(4)))),
   \end{aligned}
\end{equation} 
where $\mathcal L_1$  and $\mathcal L_2$ represent two distinct linear transformations, each with input size of 64 and output size of 64, and ${\rm ReLU}(\cdot)$ denotes  the rectified linear unit activation function.

Next, the node representation vectors undergo another linear transformation to map them into values. Following this, additional processing steps—similar to those in  \cite{b12}—are applied, including the operations described in \eqref{exp} and \eqref{proj}, giving
\begin{equation} \label{exp}
  \begin{aligned}
\mathbf{y}_2=\exp (-{\rm SoftPlus}({\mathcal L}_3 (\mathbf{Y}_1)+6),
   \end{aligned}
\end{equation} 
where ${\rm SoftPlus}(\cdot)$ denotes the Softplus activation function, and ${\mathcal L}_3$ is a linear function that maps a vector to a scalar value. 
% To extract the power control coefficients for the active users while
To ensure that the results satisfy the constraints in \eqref{cons},  projection operation ${\rm Proj}_{\mathcal{S}}(\cdot)$ in \cite{b10} are applied, yielding the power control coefficients as
\begin{equation}  \label{proj}
  \begin{aligned}
  \hat{\mathbf{M}}={\rm Proj}_{\mathcal{S}}({\mathbf Y}_2 ),
   \end{aligned}
\end{equation} 
where $\mathbf{Y}_2$ is obtained by reshaping the vector $\mathbf{y}_2$ into a matrix of size $M \times K$.
% , and the diagonal matrix $\boldsymbol{\Phi}_{\rm mask}$ shares the same diagonal elements as $\boldsymbol{\Phi}$.
\subsection{Attention Mechanism  for Graph Neural Network} \label{atten}
  Considering all layers of the attention mechanism share the same structure, differing only in the sizes of their input and output,   the following description uses the $t$th layer as a representative example.
 Denoting the  representation vector of the $i$th node as input in the $t$th layer as $\mathbf{x}_i (t-1)$,  the output node representation vector is calculated by
 \begin{equation}
  \begin{aligned}
\mathbf{x}_i(t)=\mathrm{ Norm}(\mathrm {ReLU}( \mathbf{y}_{i}^{\rm AP}(t) + \mathbf{y}_{i}^{\rm UE}(t))), 
\end{aligned}
\end{equation}
where $\mathrm{ Norm}(\cdot)$ denotes the normalization operation. Here $\mathbf{y}_{i}^{\rm AP}(t)$ represents the component of the update through the attention mechanism, based on the information of nodes that are associated with the same APs as the $i$th node, and it can be expressed as
\begin{equation}\label{17}
  \begin{aligned}
&\mathbf{y}_{i}^{\rm AP}(t)=\mathcal{L}^{{\rm AP}, t}_1 \left( \mathbf{x}_i(t-1)\right) + \\
&\mathcal{L}^{{\rm AP}, t}_5 \left(\sum \limits_{j \in \bm {\mathcal{V}}_{i}^{\rm AP}}  \alpha_{ i, j}^{\rm AP} \left[ \mathcal{L}^{{\rm AP},t}_2 \left( \mathbf{x}_j(t-1)\right)
            + \mathcal{L}_{{\varphi}}^t \left( \varphi_{i,j}\right) \right] \right),
\end{aligned}
\end{equation}
  \begin{equation}
  \begin{aligned}
&\alpha_{i, j}^{\rm AP}= \\
&\frac{\left< {\mathcal{L}^{{\rm AP},t}_3\left( \mathbf{x}_i(t-1)\right)},       \mathcal{L}^{{\rm AP},t}_4\left( \mathbf{x}_j(t-1) \right)+ \mathcal{L}_{{\varphi}}^t\left( \varphi_{i,j}\right) \right>}{\sum \limits_{u \in {\bm{\mathcal{V}}_{i}^{\rm AP}}}\left<{\mathcal{L}^{{\rm AP},t}_3\left( \mathbf{x}_i(t-1)\right)},  \mathcal{L}^{{\rm AP},t}_4\left( \mathbf{x}_u(t-1) \right)+ \mathcal{L}_{{\varphi}}^t\left( \varphi_{i,u} \right)  \right>},
\end{aligned}
\end{equation}
where  $\mathcal{L}^{{\rm AP}, t}_q$ ($q=1,\cdots,5$) denotes $5$  linear transformations for capturing the information from AP-type edges in the $t$th layer, while $ \mathcal{L}_{{\varphi}}^t$ is a linear transformation used to capture pilot contamination information in the $t$th layer.
Moreover, ${\bm {\mathcal{V}}}_{i}^{\rm AP}$ denotes the set of nodes that share the same AP as the $i$th node, $\left<\mathbf{a}, \mathbf{b}\right>=  \exp({\mathbf{a}^{\rm T} \mathbf{b}}/{\sqrt{D}})$, and $D$ is the length of  $\mathbf{a}$.

Note that $\mathbf{y}_{i}^{\rm UE}(t)$ is obtained through a similar process, but without considering pilot contamination. Specifically, $\mathcal{L}_{{\varphi}}^t\left( \cdot\right)$ is removed, as pilot contamination only arises between users.

\section{Numerical Results}
\subsection{Simulation Setup}
In this simulation,  the \makebox{large-scale} fading coefficient $\beta_{m,k}$ in $\rm dB$  is modeled as
  \begin{equation}
  \begin{aligned}
\beta_{m,k} = \mathrm{PL}_{m,k} + z_{m,k},     %\: [\mathrm{dB}]
  \end{aligned}
\end{equation}
where $\mathrm{PL}_{m,k}$ denotes the path loss in $\rm dB$, modeled using a \makebox{three-slope} model, and $z_{m,k}$ accounts for shadow fading, following  $\mathcal{CN}(0, \sigma_{\rm sh}^2)$. The same large-scale fading parameters  as in \cite{b14} are used to characterize the propagation path. The noise signal is simulated using the same parameters as in \cite{b12}.

The lengths for the coherence block and pilot sequence are set to $T_{\rm c} = 200$ and $T_{\rm p} =18$, respectively.  A total of $T_{\rm p}$ orthogonal pilot sequences are allocated among UEs. The first $K$ sequences  (corresponding to the actual number of UEs) are assigned, while for cases where $K>T_{\rm P}$,  the remaining \makebox{$K- T_{\rm p}$} sequences are randomly selected from  the $T_{\rm P}$  available sequences.
Simulation experiments are conducted in a CFmMIMO system with MMSE channel estimation in \eqref{3} and conjugate beamforming in \eqref{e1}.

\subsection{Dataset Preparation and Padding Strategy}
To address the requirement that all tensors within a batch must have the same shape in modern deep-learning frameworks such as PyTorch, which is used in our implementation, we employ a zero-padding strategy to handle training datasets with a varying number of UEs.
Specifically, $K$ is set as the maximum number of UEs, $K_{\rm max}$, within the coverage area. If the actual number of  UEs, $K_{\rm act}$, is smaller than $K_{\rm max}$, an extra $K_{\rm max}-K_{\rm act}$ UEs are added as padding.
% To differentiate between actual and padded UEs, each node is assigned a binary attribute, ${\rm Mask}$, and  ${\rm Mask}_i=1$ indicates that the UE  associated with the $i$th node is served, while  ${\rm Mask}_i=0$ denotes a padded UE.
During training data preparation, zero-padding is applied to  $\mathbf{B}$ and $\boldsymbol{\Phi}$, producing  a new $\mathbf{B}$   of size $M \times K_{\rm max}$ and a new $\boldsymbol{\Phi}$ of size $K_{\rm max} \times K_{\rm max}$ as actual inputs, ensuring uniform input dimensions across training batches.
% For simplicity of notation,  $\mathbf{B}$  and $\boldsymbol{\Phi}$ are used to refer to their padded counterparts, and $K=K_{\rm max}$   in the following discussion unless explicitly stated.
Accordingly, the following modifications are made to eliminate the influence of  padded UEs: in \eqref{proj}, ${\mathbf Y}_2$ is replaced with  ${\mathbf Y}_2 \boldsymbol{\Phi}_{\rm mask}$; 
in \eqref{17}, $\alpha_{i, j}^{\rm AP}$  is substituted with  $\varphi_{j,j} \alpha_{i, j}^{\rm AP}$.
Here, the diagonal matrix $\boldsymbol{\Phi}_{\rm mask}$ shares the same diagonal elements as the padded $\boldsymbol{\Phi}$.

\subsection{Methods Compared and Performance Metrics}
We compared the performance of the proposed method with two other power allocation methods: the APG method and the transformer-based method\footnote{The proposed network and transformer-based network were trained on the NVIDIA H200 GPU.}.
The APG method, a traditional optimization-based method as described in \cite{b10}, serves as a baseline to evaluate the performance of the proposed graph attention network, with the smoothness parameter $\lambda=3$ in \eqref{u}. 
The transformer-based method of \cite{b12}, a learning-based method also referred to as PAPC, is included for comparison as well.
Considering that all existing GNN-based methods thus far are under the assumption of free pilot contamination, whereas our proposed method highlights its ability to handle pilot contamination, GNNs are not included for comparison.

The empirical cumulative distribution function (CDF) of the \makebox{per-UE} SE serves as a tool to visualize the SE distribution. Besides, we provide the running times of methods to assess their feasibility in real-time applications.
\subsection{Performance Comparison in Different Scenarios}
 In this section, we compare the performance of $3$ methods  across $5$ scenarios, demonstrating the adaptability and scalability of the proposed method. 
The parameter settings for $5$  scenarios are listed in \makebox{Table \ref{tab}}. For the first scenario, as an example,  $K_{\rm  max}=8$ UEs are served by $M=16$ APs, with both UEs and APs uniformly randomly distributed within a coverage area of $0.16$ ${\rm km}^2$. Since $K<T_{\rm p}$ in this case, pilot contamination does not occur. The number of UEs is not varying, thus $K=K_{\rm  max}=K_{\rm  min}$ for all sample data. If  the number of UEs is varying, as in Scenarios 3 and 5,  the number of UEs  changes within $\left[K_{\rm  min}, K_{\rm  max}\right]$. In Scenarios 2-5 with $K_{\rm  max}>T_{\rm p}$, pilot contamination arises accordingly.   All scenario results are tested on $500$ samples, and the number of training samples for each scenario is provided in \makebox{Table \ref{tab}}.
\begin{table}[t]
\centering
\caption{Parameters for $5$ Scenarios}
\label{tab}
\setlength{\tabcolsep}{4.2pt} % Reduce column padding
\setlength{\extrarowheight}{1.7pt}
\begin{tabular}{|c|c|c|c|c|c|c|}
\hline
\textbf{Parameters} & \textbf{\makecell{Scen. 1 }} & \textbf{\makecell{Scen. 2}} & \textbf{\makecell{Scen. 3}} & \textbf{\makecell{Scen. 4}} & \textbf{\makecell{Scen. 5}} \\ \hline
\makecell{Coverage Area (${\rm km}^2$)} &0.16 &0.32 &0.32 &0.32 &0.32 \\ \hline
%\makecell{AP Density} & 100 & 100 & 100 & 200 & 200 \\ \hline
M& 16  & 32  & 32 & 64 & 64 \\ \hline
$K_{\rm max}$  & 8& 20 & 20& 40 & 40 \\ \hline
$K_{\rm min}$  & 8& 20 & 10& 40 & 20 \\ \hline
\makecell{Pilot Contamination}& $\times $& $\checkmark$ & $\checkmark$& $\checkmark$ & $\checkmark$ \\ \hline
\makecell{Varying UEs} & $\times$ &$\times$ & $\checkmark$&$\times$& $\checkmark$ \\ \hline
\makecell{Training Samples} & 50K &100K & 800K &100K & 800K\\ \hline
%\makecell{Test \\Samples} & 200 &200 & 200 &200 & 200 \\ \hline
\end{tabular}
\end{table}

\begin{figure}[t]
  \centerline{\includegraphics[width=8cm,height=5.59cm]{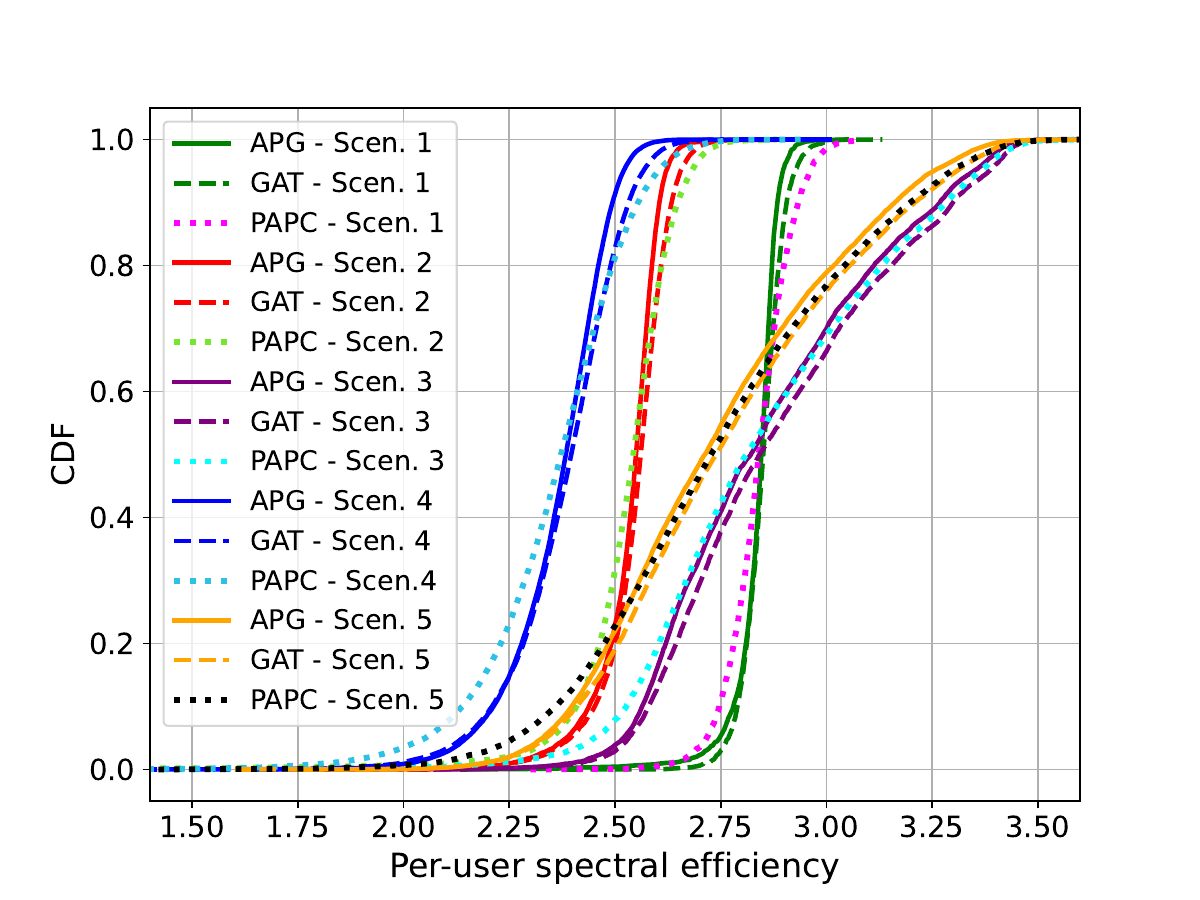}}
%  \vspace{1.5cm}
\caption{Result comparison between $3$ methods across $5$ scenarios.}
\label{results}
\end{figure}
\begin{figure}[t]
  \centerline{\includegraphics[width=8cm,height=5.3cm]{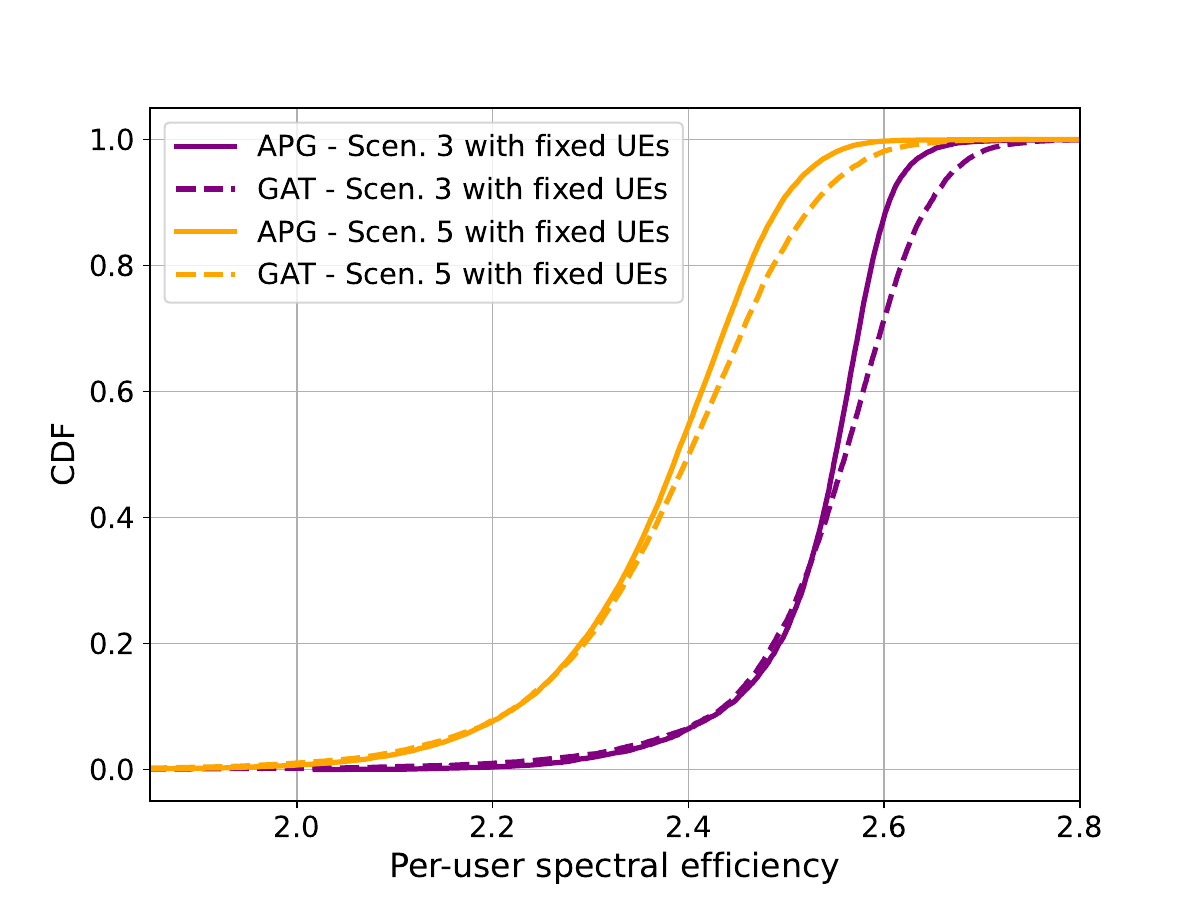}}
%  \vspace{1.5cm}
\caption{Performance comparison for the proposed method trained with a varying number of UEs but tested with  fixed UEs.}
\label{results2}
\end{figure}

The CDF results for $5$ scenarios are presented in \makebox{Fig. \ref{results}}, and the quantitative results corresponding to Eq.\eqref{u} is provided in \makebox{Table \ref{22}}.
These values in \makebox{Table \ref{22}} represent the average results over all test samples.
Note that the proposed  Graph ATtention network method is labeled as GAT in  Figs. \ref{results} and \ref{results2}  for simplicity.
In the context of the max-min fairness objective, a steeper CDF curve that shifts towards the right indicates a more desirable result, as it suggests improved fairness across UEs, as well as higher SEs for all UEs. 
Observing \makebox{Fig. \ref{results}} and \makebox{Table \ref{22}}, it can be concluded that the proposed method outperforms  both APG and PAPC in all scenarios, including variations in communication network size, pilot contamination, and whether the number of UEs is fixed or not.
Note that in Scenarios 3 and 5, the curves do not ascend as sharply as those of the other three scenarios. This is because the number of active UEs varies for each test sample, leading to a wider range of SE per UE.
\begin{table}[ht]
\centering
\caption{Quantitative results showing the corresponding values  of \eqref{u} (3 methods, $5$ scenarios).}
\label{22}
\begin{tabular}{c c c c}
\hline
\textbf{Scenarios} & \textbf{APG} & \textbf{ PAPAC} & \textbf{Proposed} \\
\hline
Scen. 1 & 1.9647 & 1.9641 & 1.9731\\
Scen. 2 & 1.7524 & 1.7363 & 1.7619 \\
Scen. 3 & 2.0219 & 2.0132 & 2.0430 \\
Scen. 4 & 1.6311 & 1.6091 & 1.6380 \\
Scen. 5 & 1.9327 & 1.9314 & 1.9541 \\
\hline
\end{tabular}
% \label{22}
\end{table}

To evaluate the flexibility of the proposed method, the trained models from Scenarios 3 and 5, which  were trained with varying numbers of UEs, are tested on fixed UE settings, i.e., Scenario 3  with $K=20$ and Scenario 5 with $K=40$. The results, presented  in \makebox{Fig. \ref{results2}}, show that the models trained on datasets with varying numbers of UEs  still demonstrate strong performance,  even surpassing the APG method.

Table \ref{tab2} presents the runtime for each method across different scenarios.  The CPU measurements are obtained on a \makebox{$64$-bit} Windows system with $16$ GB RAM and an Intel Core i5 ($2.60$ GHz), while the GPU measurements are obtained using an NVIDIA A40.  
As shown, compared to $5.5185 \, [{\rm s}]$ GPU processing time for  APG in Scenario 5 with $64$ APs and $40$ UEs,  the proposed method  achieves a processing time of  \makebox{$2.2\, [{\rm ms}]$} on the GPU despite $592\, [{\rm ms}]$ on the CPU, highlighting the \makebox{real-time} performance of the learning-based method and demonstrating its potential for power control in \makebox{real-time} applications. 
Note that although the PAPC's is  slightly less than the proposed model, the performance improvement justifies the negligible time difference.

\begin{figure}[htbp]
    \centering
    \includegraphics[width=8cm,height=6cm]{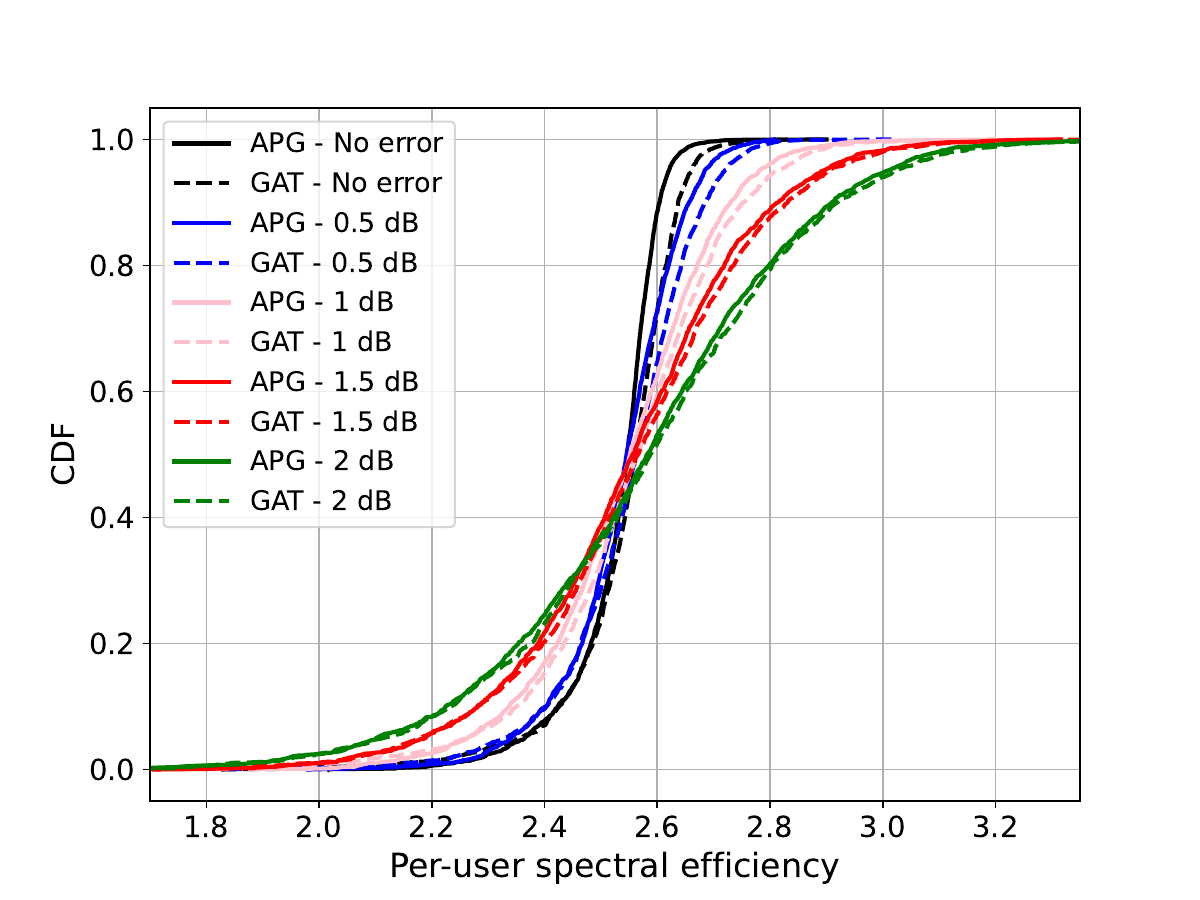}
     \caption{Performance comparison for different estimation errors in Scen. 2}   
     \label{Err}
\end{figure}

While TDD systems leverage channel reciprocity, practical implementations can encounter errors due to hardware impairments or calibration mismatches, leading to  large-scale channel estimation errors.
To assess the impact of large-scale channel estimation errors on the performance of the proposed model, another experiment was conducted, with the result is presented in Fig. \ref{Err}, and the experimental settings are as follows. 
Let $\beta_{m,k}^{[\rm dB]}= 10\log_{10} (\beta_{m,k})$ denote the large-scale fading coefficient in the decibel scale.
To model the effect of estimation errors in the simulations, we add additive white Gaussian noise to the channel coefficients, which allows us to realistically capture the impact of imperfect channel knowledge on system performance \cite{Li_Residual, Zeng_Joint, Peng_Robust}. 
The corresponding estimated values with errors can then be modeled  as 
\begin{equation*}
  \begin{aligned}
 % \color{blue}
\hat{\beta}_{m,k}^{[\rm dB]} =  \beta_{m,k}^{[\rm dB]}+\mathcal{CN}(0,\sigma^2).
\end{aligned}
\end{equation*}
To evaluate the impacts of estimation errors, we consider different values of  $\sigma$: $0.5,1,1.5$, and $2$. The experiments were conducted under Scenario 2, and the resulting  CDFs are given in Fig. \ref{Err}.
As shown in Fig. \ref{Err}, the performance degrades with increasing $\sigma$, as reflected by the CDF results. However, the proposed model consistently achieves nearly the same performance as the APG algorithm, even when $\sigma$ is as high as 2, which validates the effectiveness and robustness of our proposed method.

\begin{figure}[t]
  \centerline{\includegraphics[width=8cm,height=5.59cm]{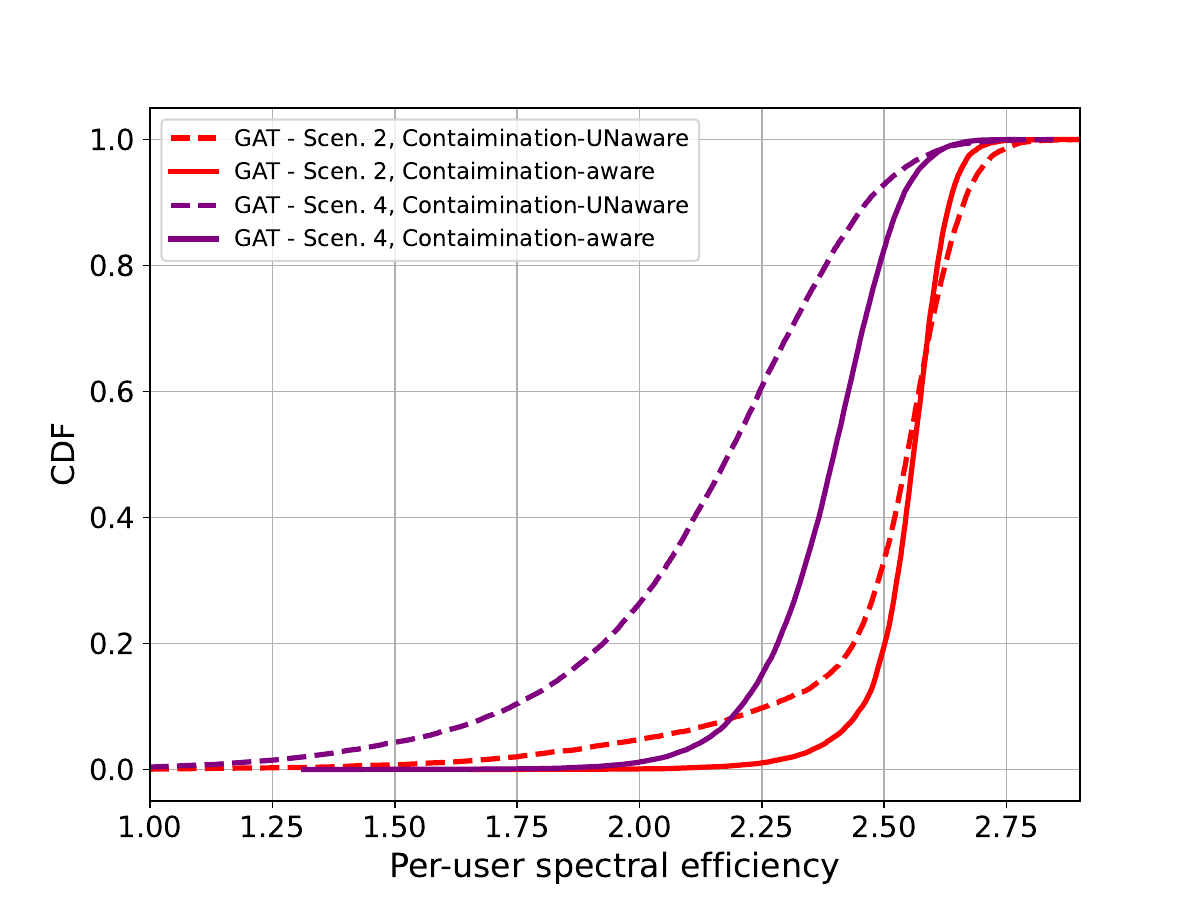}}
%  \vspace{1.5cm}
\caption{Ablation study on pilot information awareness.}
\label{Ablation}
\end{figure}

Next, to assess the performance gain introduced by incorporating pilot contamination information, an ablation study is conducted by removing the pilot-related input, $\Phi$, from the proposed model. Using Scenarios 2 and 4 as representative examples, the resulting CDFs of the per-user spectral efficiency are illustrated in Fig.~\ref{Ablation}.
As observed, the CDF curves of the GAT model with pilot contamination awareness (in solid line) in both scenarios are significantly shifted to the right compared to its counterpart without pilot information (in dashed line).
This clearly demonstrates the importance of including pilot contamination information in the GNN input, highlighting its critical role in enhancing power control performance in CF-mMIMO systems.

In conclusion, the proposed method achieves performance comparable to APG.
Moreover, it significantly reduces computational complexity, highlighting the superior suitability of GNNs for CFmMIMO applications.

\begin{table}[htbp]
\centering
\caption{Runtime comparison between methods (in seconds).}
\label{tab2}
\setlength{\tabcolsep}{4.2pt} % Reduce column padding
\setlength{\extrarowheight}{1.5pt}
\begin{tabular}{|c|c|c|c|c|c|c|}
\hline
\textbf{} & \multicolumn{2}{c|}{\textbf{Scen. 1}} & \multicolumn{2}{c|}{\textbf{Scen. 3}} & \multicolumn{2}{c|}{\textbf{Scen. 5}} \\
\cline{2-7} % This line spans from column 2 to column 7
& \makecell{CPU } & \makecell{GPU } & \makecell{CPU } & \makecell{GPU } & \makecell{CPU} & \makecell{GPU} \\
\hline
APG & 1.5626 & 1.1391& 3.5581 & 2.4226 & 11.2485 & 5.5185 \\ % Assuming APG only has CPU
\hline
PAPC & 0.0022 & 0.0012 & 0.0134 & 0.0013 & 0.0167 & 0.0014 \\
\hline
Proposed & 0.0083 & 0.0020 & 0.0613 & 0.0021 & 0.5920 & 0.0022 \\
\hline
\end{tabular}
\end{table}

\section{Conclusion}
A graph attention method has been proposed for downlink power control in CFmMIMO systems. 
The proposed method not only effectively managed power control in the presence of pilot contamination but also demonstrated strong adaptability to varying communication network sizes and both fixed and dynamic numbers of UEs. Additionally, the neural network was trained in a self-supervised manner, eliminating the need for the labor-intensive process of labeling the large volume of samples. Experimental results demonstrated that the proposed method outperforms APG and transformer-based methods in terms of CDF for SE across various scenarios.

\end{document}